\pdfoutput=1
\documentclass[11pt]{article}

\usepackage[preprint]{acl}
\usepackage{times}
\usepackage{latexsym}
\usepackage{subfigure}
\usepackage{amsmath,amssymb,amsfonts}
\usepackage{algorithmic}
\usepackage{multirow}
\usepackage{array}
\usepackage[T1]{fontenc}
\usepackage[utf8]{inputenc}
\usepackage{microtype}
\usepackage{inconsolata}
\usepackage{marvosym}
\usepackage{graphicx}

\author{Jingwang Huang\textsuperscript{1}, \ Jiang Zhong\textsuperscript{1}\thanks{Corresponding author}, \ Qin Lei\textsuperscript{1}, \ Jingpeng Gao\textsuperscript{1}, \ Yuming Yang\textsuperscript{1}\\
\ \textbf{Sirui Wang\textsuperscript{2,3}}, \ \textbf{Peiguang Li\textsuperscript{3}} \ \textbf{Kaiwen Wei\textsuperscript{1*}} \\
  \textsuperscript{1} College of Computer Science, Chongqing University, China \\
  \textsuperscript{2} Department of Automation, Tsinghua University China \\
  \textsuperscript{3} Meituan Inc., Beijing, China \\
  \texttt{huangjingwang@stu.cqu.edu.cn},  
  \texttt{\{zhongjiang, weikaiwen\}@cqu.edu.cn} \\
  } 
\title{Latent Distribution Decoupling: A Probabilistic Framework for Uncertainty-Aware Multimodal Emotion Recognition}

\begin{document}
\maketitle
\begin{abstract}
Multimodal multi-label emotion recognition (MMER) aims to identify the concurrent presence of multiple emotions in multimodal data. Existing studies primarily focus on improving fusion strategies and modeling modality-to-label dependencies. 
However, they often overlook the impact of \textbf{aleatoric uncertainty}, which is the inherent noise in the multimodal data and hinders the effectiveness of modality fusion by introducing ambiguity into feature representations.
To address this issue and effectively model aleatoric uncertainty, this paper proposes Latent emotional Distribution Decomposition with Uncertainty perception (LDDU) framework from a novel perspective of latent emotional space probabilistic modeling. Specifically, we introduce a contrastive disentangled distribution mechanism within the emotion space to model the multimodal data, allowing for the extraction of semantic features and uncertainty. Furthermore, we design an uncertainty-aware fusion multimodal method that accounts for the dispersed distribution of uncertainty and integrates distribution information. Experimental results show that LDDU achieves state-of-the-art performance on the CMU-MOSEI and M$^3$ED datasets, highlighting the importance of uncertainty modeling in MMER. Code is available at \url{https://github.com/201983290498/lddu\_mmer.git}.
\end{abstract}
\section{Introduction}
Human interactions convey multiple emotions through various channels: micro-expressions, vocal intonations, and text. 
Multimodal multi-label emotion recognition (MMER) seeks to identify multiple emotions (e.g., happiness, sadness) from multimodal data (e.g., audio, text, and video) \citep{b3}. It could support many downstream applications such as emotion analysis \cite{b1}, human-computer interaction \cite{b2}, and dialogue systems \cite{dialoguegcn}.  
\begin{figure}[t]
\small
\centerline{\includegraphics[width=0.48\textwidth]{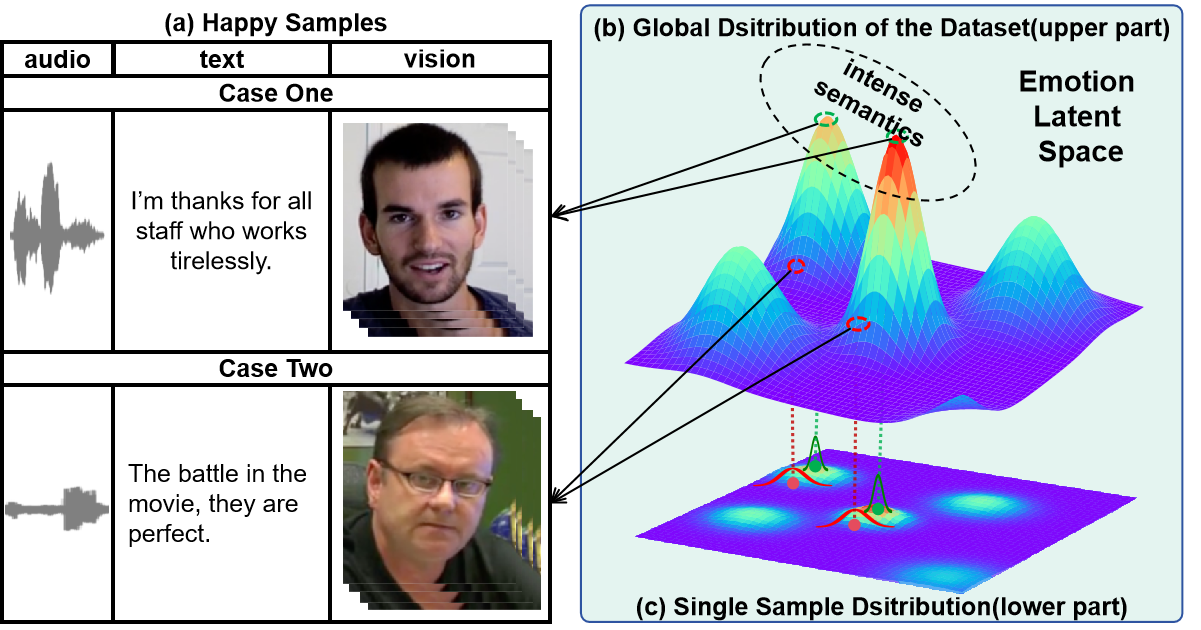}}
\caption{An illustration of aleatoric uncertainty in MMER task. When adopting Gaussian distribution modeling in latent emotion space, case two's semantic feature is more fuzzed with a larger variance due to cold speaking style than case one. Meanwhile, case one has stronger emotion intense located closer to the center of global distribution. }
\label{intro_sample}
\end{figure}

The main topics of MMER lie in extracting emotion-relevant features by effectively fusing multimodal data and modeling modality-to-label dependencies \citep{b3, b23}. To implement the fusion of multimodal data, some work \cite{b9,b17} employed projection layers to mitigate the modality gap \citep{clip}, while other methods \cite{b3,b1} utilized attention mechanisms. Additionally, several studies \cite{b23,b7,b6,b26} decomposed modality features into common and private components. Recently, CARET \citep{b5} introduced emotion space modeling, where emotion-related features were extracted prior to fusion, achieving state-of-the-art performance. Regarding modality-to-label dependencies, many approaches \cite{b3,b7} leveraged Transformer decoders to capture the relationships between label semantic features and fused modality sequence features.

However, these approaches primarily focus on semantic features while overlooking \textbf{aleatoric uncertainty} \citep{uncertainty}, which represents inherent noise in the data and is commonly modeled using multivariate Gaussian distributions \cite{do2008multivariate} (for a detailed background, please refer to Appendix~\ref{sec:appendixE}). In the context of MMER, such uncertainty primarily arises from factors such as personalized expressions, variations in emotional intensity, and the blending of coexisting emotions \citep{b4}.
For instance, as illustrated in Fig.~\ref{intro_sample}, from a macroscopic perspective, both samples convey happiness, yet case one exhibits more pronounced facial expressions compared to case two. From a distributional perspective, case one demonstrates more concentrated semantic features near the center of the dataset’s overall distribution, whereas case two presents features with greater variance, positioned farther from the center. This aleatoric uncertainty introduces ambiguity into semantic feature representations, thereby diminishing the effectiveness of modality fusion in existing MMER approaches \citep{b62}.

To model aleatoric uncertainty in MMER, several challenges need to be addressed:
(1) \textit{How to represent aleatoric uncertainty:} Emotional cues are embedded in multimodal sequences, with each modality contributing differently to emotion expression, making it difficult to extract and disentangle emotional features. When modeled with multivariate Gaussian distributions, samples with the same label often cluster together despite semantic fuzziness. An effective distribution must capture both the central tendency and calibrate variance of emotional features, which is particularly challenging.
(2) \textit{How to integrate semantic features with aleatoric uncertainty:} Higher uncertainty leads to more dispersed distributions, complicating emotion recognition. Without calibrated uncertainty, semantic features can become ambiguous and less informative. Thus, effective strategies for calibrating and integrating uncertainty are crucial to ensure robust and discriminative emotion representations.

To address these challenges, we propose Latent Distribution Decouple for Uncertainty-Aware MMER (LDDU) from the perspective of latent emotional space probabilistic modeling. For the first challenge, to represent aleatoric uncertainty, LDDU extracts modality-related features using Q-Former-like alignment \citep{b45}. We then design a distribution decoupling mechanism based on Gaussian distributions to model uncertainty. To further enhance the distinguishability of these distributions, contrastive learning \citep{chen2020simple} is employed. For the second challenge, to effectively integrate the distributional information, we draw inspiration from uncertainty learning \citep{b28,b29,b26} and develop an uncertainty-aware fusion module, which is accompanied by uncertainty calibration. Experimental results on the CMU-MOSEI and M$^3$ED datasets show that LDDU achieves state-of-the-art performance. Specially, it surpasses strong baseline CARAT 4.3\% miF1 on CMU-MOSEI under unaligned settings.
In summary, the contributions of this work are as follows: 
\begin{itemize} 
\item We introduce latent emotional space probabilistic modeling for MMER. To the best of our knowledge, this is the first work to leverage emotion space distribution for capturing aleatoric uncertainty in MMER.
\item We propose LDDU, which models the emotion space to extract emotion features, then uses contrastive disentagled learning to represent latent distributions and recognizes emotions by integrating both semantic features and calibrated uncertainty.
\item Experiments on CMU-MOSEI and M$^3$ED datasets demonstrate that the proposed LDDU method achieves state-of-the-art performance, with mi-F1 improved 4.3\% on the CMU-MOSEI unaligned data. \end{itemize}

\begin{figure*}[htbp]
\small
\centerline{\includegraphics[width=1\textwidth]{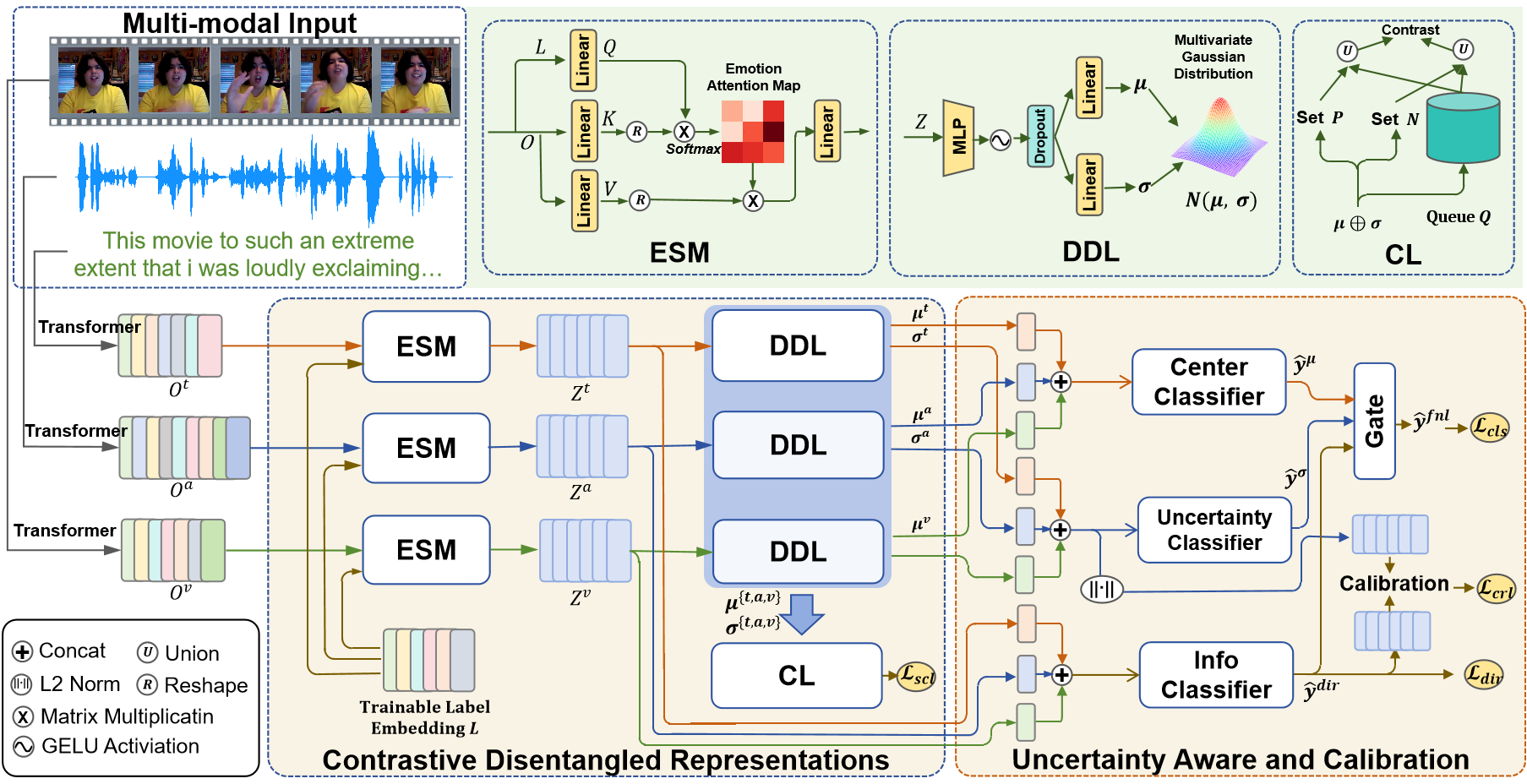}} 
\caption{The proposed LDDU framework consists of three components: (1) the transformer-base unimodal extractor (2) a contrastive learning-based emotion space decomposition module , and (3) an uncertainty-aware fusion and uncertainty calibration module.}
\label{framework}
\end{figure*}
\section{Related Work}
\textbf{Multimodal Multi-label Emotion Regression.} It aims to infer human emotions from textual, audio, and visual sequences in video clips, often encompassing multiple affective states. The primary challenges in MMER is integrating multimodal data. Early studies like MISA \citep{b23} address modality heterogeneity by decoupled invariant and modality-specific features for fusion. MMS2S \citep{b9} and HHMPN \citep{b3} focused on modeling label-to-label and label-to-modality dependencies using Transformer and GNNs network. Recent approaches \citep{b5, b6, b7} incorporated advanced training techniques; for example, TAILOR \citep{b7} utilized adversarial learning to differentiate common and private modal features, while AMP \citep{b24} employed masking and parameter perturbation to mitigate modality bias and enhance robustness. However, these works all model from multimodal fusion instead of emotion latent space. \\
\textbf{Uncertainty-aware Learning and Calibration.} Deep models often overconfidently assign high confidence to incorrect predictions, making uncertainty-aware learning essential to ensure confidence accurately reflects prediction uncertainty \citep{b28}. The primary goal is to calibrate model confidence to match the true probability of correctness. There are two main approaches: calibrated uncertainty \citep{b28} and ordinal or ranking-based uncertainty \citep{b29}. Calibration methods, such as histogram binning, temperature scaling, and accuracy versus uncertainty calibration \citep{b30,b28,b31}, align predicted confidence with actual correctness. Meanwhile, ranking-based methods like Confidence Ranking Loss (CRL) \citep{b29} enforce accurate confidence rankings among correctly classified samples based on feature distinctiveness.\\
\textbf{Uncertainty-based Multimodal Fusion.} Uncertainty learning enhances multimodal fusion across tasks. \citet{b27} employed Bayesian deep learning and AvU to guide fusion, while \citet{b26} used temporal-invariant learning to reduce redundancy and noise, improving robustness. But these methods incorporate uncertainty without calibration. In contrast, COLD \citep{b25} leveraged GURs to model feature distributions across modalities, quantifies modality contributions with variance norms, and integrated both calibrated and ranking-based uncertainty to regulate fusion variance. However, there hasn't exploration of uncertainty-aware for MMER.

\section{Methodology}
\subsection{Preliminary}
MMER is typically modeled as a multi-label task. 
Suppose ${X}^v \in \mathbb{R}^{s_v \times d_v} $, $ {X}^{a} \in \mathbb{R}^{s_a \times d_a} $ and $ {X}^{t} \in \mathbb{R}^{s_t \times d_t } $ denote the features of the text, visual, and audio modalities, respectively. In this context, $s_v$, $s_a$, $s_t$ denote the length of the feature sequences, while $d_v$, $d_a$,$d_t$ is the dimension of each features sequence. Given a multimodal sequential dataset in joint feature space $ \mathcal{X}^{v, a, t} $,  denoted as $\mathcal{D} = \{ (X_i^v, X_i^a, X_i^t, y_i)\}_{i=1}^{N} $, the objective of the MMER is to learn the function $\mathcal{F}$: $\mathcal{D} \rightarrow \mathcal{Y}$. Here, $N$ is the size of dataset $ \mathcal{D} $ and $X_i^v, X_i^a, X_i^t$ represent the visual, audio and textual sequences of the $i$-th sample. $ \mathcal{Y} \in \mathbb{R}^{q} $ represent the emotion space containing $q$ multiple coexisting emotion labels. 

In this section, we describe LDDU framework, which comprises three components (in Figure \ref{framework}).
\subsection{Uni-Modal Feature  Extractor}
Follow the work of \citet{b5}, we conduct experiments on CMU-MOSEI \citep{b19} and M3ED \citep{b16} datasets. 
In these two benchmark, facial keypoints $X^v$ via the MTCNN algorithm \citep{b32}, acoustic features $X^a$ with Covarep \citep{b33} and text features $X^t$ of sample $X$ are extracted using BERT \citep{b20}. To capture content sequence dependencies, we employ $n_v$, $n_a$, and $n_t$ Transformer layers as unimodal extractors, generating modality visual features $O^v \in \mathbb{R}^{s_v \times d_v}$, audio features $O^a \in \mathbb{R}^{s_a \times d_a}$, and text features $O^t \in \mathbb{R}^{s_t \times d_t}$. Each modality $O^m$ is derived from its sequence data $[o_1^m, \dots, o_{s_m}^m]$, $m \in \{v, a, t\}$.

\subsection{Contrastive Disentangled Representation}

\subsubsection{Emotion Space Modeling}
A primary challenge in emotion space modeling is the establishing emotion representations within a unified joint embedding space. Inspired by the Q-Former’s structure \citep{b45}, we introduce trainable emotion embeddings $L = [l_1, l_2, ..., l_{q}]$, where each $l_i$ represents an emotion and $q$ is the number of label. Because emotion-related cues may be distributed across different segments of the sequential data, we employ an attention mechanism to automatically extract relevant features for each emotion. Since modality-related features $O^m$ and $L$ reside in different feature spaces, we use projection layers to compute the similarity $a^m_{ij}$ between frame's feature $o^m_j$ and the label $l_i$. After obtaining the similarity matrix $A^m=\{a^m_{ij}\}$, $Y^m$ is projected to extract modality-specific label-related features $Z^m \in \mathbb{R}^{q \times d_h}$, where $ d_h $ is the  dimension of modality-specific label-related features. This process could formalized as follows:
\begin{equation}
a^m_{ij} = \frac{exp(Proj(l_i)^{T}Proj(o^m_j))}{\sum^{s_m}_{j'=1} exp(Proj(l_i)^{T} Proj(o^m_{j'}))}   \label{eq:21} \\    
\end{equation}
\begin{equation}
Z^m =  Linear(A^m Proj(O^m)) \label{eq:22}
\end{equation}
where $Proj$ represents the projection layer.

To facilitate the learning of emotion representations $L$, we concatenate the multimodal features of $i$-th sample into $F_{dir} = [Z_i^v, Z_i^a, Z_i^t]$ and process them with an MLP-based info classifier employing sigmoid activation functions to generate the final prediction $\hat{y}^{dir}_i = [\hat{y}^{dir}_{i1}, \dots, \hat{y}^{dir}_{iq}]$. The loss function $\mathcal{L}_{dir}$ is defined as follows: 
\begin{equation}
\mathcal{L}_{dir}  = \frac{1}{N} \sum_{i=1}^{N} \text{BCE}(y_i,\hat{y}^{dir}_i)\label{eq:23}
\end{equation}
where BCE(.) is the BCE loss. 
\subsubsection{Contrastive Distentangled Distribution Modeling}
\begin{figure}[t]
\centerline{\includegraphics[width=0.48\textwidth]{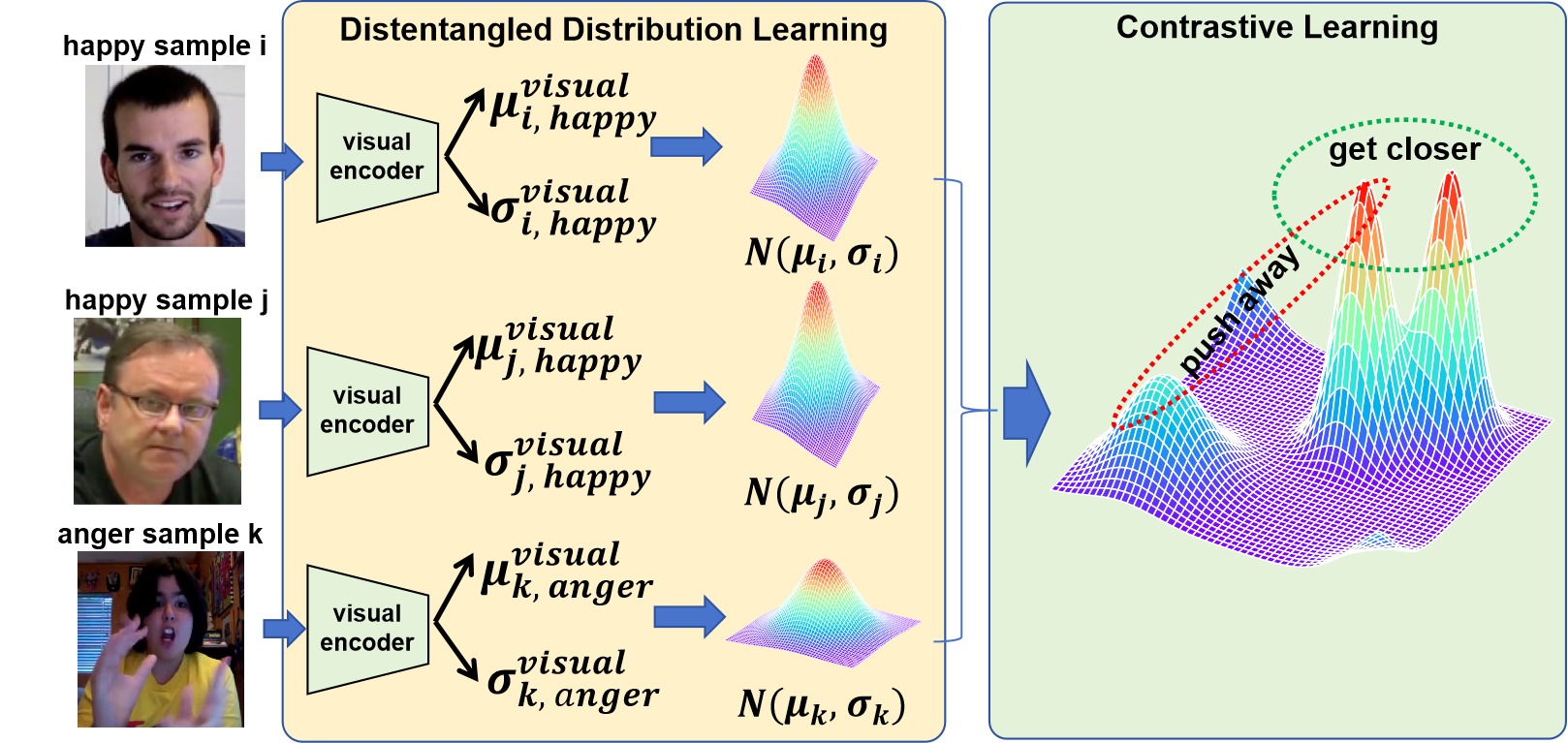}}
\caption{In the latent emotion space, we decouple emotion-related, modality-specific samples into separate distributions and use CL to group samples of the same category together while separating samples from different categories.}
\label{method2}
\end{figure}

This module is composed of distentangled distribution learning (DDL) and contrastive learning (CL). As illustrated in Figure \ref{method2}, our architecture incorporates disentangled representation learning (DRL) \citep{b17,b47} to establish label-related features into  latent probabilistic distribution in emotion space. Specifically, we model multimodal emotion representations $[Z^v, Z^a, Z^t]$ as multivariate normal distributions $\mathcal{N}$. For each label-related features $[Z_i^v, Z_i^a, Z_i^t]$, we leverage an encoder (MLP in this paper) and two fully connected layers to obtain the latent distributions $\mathcal{N}(\mu_i^v, \sigma_i^v)$, $\mathcal{N}(\mu_i^a, \sigma_i^a)$ and $\mathcal{N}(\mu_i^t, \sigma_i^t)$. where $\mu_i^t$ represents the semantic features of the text modality for emotion label $i$ \citep{b25} and $\sigma_i^t$ reflects the distribution region in latent space.

To ensure that the latent distribution $\mathcal{N}(\mu_i^m, \sigma_i^m)$ accurately captures feature differences for each label across modality $m$, we employ Contrastive Learning (CL). CL could groups similar samples together and enhances the model's ability to distinguish between different classes \citep{b48, b50}. Formally, given the variations in latent distributions across labels and modalities, we categorize them into $3q$ potential emotional distributions. For a batch of $s_B$ samples $\mathcal{B}$, each sample generates $3q$ label-related and modality-specific emotion distributions, totaling $3 \times q \times s_B$ distributions. Each distribution in $\mathcal{B}^+$ is considered a positive sample if the related sample contains the corresponding emotion. For each positive distribution $e \in \mathcal{B}^+$, we identify its positive set $\mathcal{P}_e(\mathcal{B})$ and negative set $\mathcal{N}_e(\mathcal{B})$ based on labels. 

Besides, we promote CL from the following two perspectives. First, \citet{b50} observes that a larger batch size can enhance the network abilities by providing more diverse negative samples in CL. We introduce a queue $Q$ of size $s_q$ to store the most recent $s_q$ emotion distributions. Thus, the final positive and negative sets for each emotion distribution become $\mathcal{P}_e(\mathcal{B} \cup \mathcal{Q})$ and $\mathcal{N}_e(\mathcal{B} \cup \mathcal{Q})$, respectively. Besides, similarity calculations between samples must consider both the centers and variances of the decoupled  distributions. We represent the distribution $e$ as follows:
\begin{gather}
e = \left(\frac{\mu_{e,1}}{|\mu_e|}, \dots, \frac{\mu_{e,d_h/2}}{|\mu_e|}, \frac{\sigma_{e,1}}{|\sigma_e|}, \dots, \frac{\sigma_{e,d_h/2}}{|\sigma_e|} \right)
\end{gather}

Finally, we introduce the SupCon loss \citep{b49} to for each emotion distribution:
\begin{equation}
\mathcal{L}_{scl}(e, \mathcal{B}^+)=\frac{-1}{|\mathcal{P}_e|} \sum_{e^+ \in \mathcal{P}_e} \log \frac{e^{z(e, e^+/\tau)}}{\displaystyle \sum_{e' \in \mathcal{T}_e} e^{z(e,e')/\tau}}
\end{equation}
where $\mathcal{T}_e$ = $\mathcal{P}_e \cup \mathcal{N}_e$, and $z$ is the similarity function between emotin distribution. To simplify the process, we calculated cosine similarity on the normalized distribution parameters:
\begin{equation}
z(e_1,e_2) = e_1^Te_2
\end{equation}
The final contrastive loss for the entire batch is:
\begin{equation}
\mathcal{L}_{scl} = \sum_{e \in \mathcal{B}^+} \mathcal{L}_{scl}(e, \mathcal{B}^+)
\end{equation}

\subsection{Uncertainty Aware and Calibration}
\subsubsection{Uncertainty-Awared Multimodal Fusion}
After modeling the emotional space, it's crucial to integrate latent semantic features with the distribution uncertainty information. We use variance to represent the distribution uncertainty in latent space, as it reflects the degree of dispersion and distribution region. 
Meanwhile, the center feature represents the semantic features of a sample \citep{b62, b25, b26}. We hypothesize that when a sample has high aleatoric uncertainty, its semantic features become fuzzier, and the distribution region in latent space becomes more discriminative for emotion recognition. Conversely, when aleatoric uncertainty is low, the semantic features are more discriminative, and the distribution region is narrower. Therefore, the fusion of center feature and variance should depend on the level of aleatoric uncertainty.

Firstly, we introduce the $i$-th sample's prediction $\hat{y}_i^{dir}$ of Info Classifier to quantify uncertainty. \citet{uncertainty} pointed out that aleatoric uncertainty can be measured by the prediction difficulty of the sample.  Specifically, 
if $Z_i$ correctly classified by Info Classifier while $Z_j$ is misclassified and needs to be decoupled for further classification. We infer that the $j$-th sample exhibits higher aleatoric uncertainty(i.e., is less informative). Consequently, the uncertainty can be represented as $d(\hat{y_i}^{dir}, y_i)$, where $\hat{y_i}^{dir}$ is the prediction of $Z_i$. 

Then, we integrate the distribution's information by fusing multimodal data. After decoupled, the samples are represented as latent distributions $\mathcal{N}\left(E^{v,a, t}, M^{v,a, t}\right)$ where $E^m$=$[\mu_1^m,...,\mu_q^m]$ and $M^m$=$[\sigma_1^m,...,\sigma_q^m]$ for each modality m. Since $E^{v,a,t}$ and $M^{v,a,t}$ have different semantics, we implement late fusion using gate network. Operationally, $(E^v$,$E^a$,$E^t)$ and $(M^v$,$M^a$,$M^t)$ are concatenated and passed through final classifier to obtain the predictions $\hat{y_i}^{\mu}$ and $\hat{y_i}^{\sigma}$. Semantic mean vector and the variance are dynamically fused according to uncertainty score:
\begin{equation}
\hat{y}_i^{fnl} = d(\hat{y}_i^{dir},y_i)\hat{y}_i^{\mu} + (1-d(\hat{y}_i^{dir},y_i))\hat{y}_i^{\sigma}
\end{equation}

For a batch of data with size $s_B$, the loss function is as follows:
\begin{equation}
\mathcal{L}_{cls} = \frac{1}{|\mathcal{B}|} \sum_{i \in \mathcal{B}} \text{BCE}(\hat{y}_i^{fnl}, y_i)
\end{equation}
\subsubsection{Uncertainty Calibration}

In this section, we impose ordinality constraint \citep{b29} to model the relationship between uncertainty and distribution variance. When well-calibrated, the uncertainty score acts as a proxy for the correctness likelihood of its prediction for the latent distribution. In other words, well-calibrated uncertainty indicates the expected estimation error, i.e., how far the predicted emotion is expected to lie from its ground truth.


It has been confirmed that: frequently forgotten samples are harder to classify, while easier samples are learned earlier in training \citep{b53, b54}. As a result, to represent the correctness likelihood values, we use the proportion of samples $r_i$ correctly predicted by the Info Classifier during the SGD  process \citep{b52, b26}. 

In our approach, the variance $\sigma_i = (\sigma_i^v, \sigma_i^a, \sigma_i^t)$ and the prediction error $d(\hat{y}_i^{dir}, y_i)$ from the Info Classifier are strongly correlated with the correctness likelihood values of emotion classification. Thus, the calibration can be formulated as follows:
\begin{align}
\scriptstyle \arg\max \ \text{Corr}(rk(\frac{1}{\|\sigma_i\|_2}, \frac{1}{\|\sigma_j\|_2}), rk(r_i,r_j)) \label{eq:31} \\
\scriptstyle \arg\max \ \text{Corr}(rk(1-d_i, 1-d_j), rk(r_i,r_j))  \label{eq:32}
\end{align}
where $rk$ is ranking and $Corr$ is correlation. When the sample contain high uncertainty, the latent distribution variance $\sigma_i$ and the prediction error $d_i=d(\hat{y}_i^{dir}, y_i)$ tend to be large, while $r_i$ tend to be small. Conversely, when the uncertainty is small, these features are reversed.

For a batch of size $s_B$, we we compute the variance norm $S$, distance vector $D$, and proxy vector $R$ for each sample:
\begin{align}
S=[&\frac{1}{||\sigma_1||_2},\frac{1}{||\sigma_2||_2},..,\frac{1}{||\sigma_{s_B}||_2}] \\
D=[&1-d_1,1-d_2,..,1-d_{s_B}] \\
&R=[r_1,r_2,..,r_{s_B}]    
\end{align}

In order to establish the ranking constraints among $S$, $D$ and $R$, we impose ordinality constraints based on soft-ranking \citep{b25, b55}. Our method employs bidirectional KL divergence to assess mismatching between the softmax distributions of pairs $(S,R)$ and $(D, R)$. Consequently, ordinality calibration loss $\mathcal{L}_{ocl}$ can be calculated as follows:
\begin{align}
\mathcal{L}_{ocl} = & KL(P_D||P_R) + KL(P_R||P_D) \notag \\ &+ KL(P_S||P_R) + KL(P_R||P_S)
\end{align}
where $P_D$, $P_R$, and $P_S$ represent the softmax distributions of features $S$, $R$, and $D$, respectively.

Overall, in the whole training process, the training loss of LDDU is as follows:
\begin{equation}
    \mathcal{L}_{total} = \mathcal{L}_{cls} + \lambda \mathcal{L}_{ocl} + \beta \mathcal{L}_{scl} + \gamma \mathcal{L}_{dir}
\end{equation}
where $\lambda$, $\beta$, and $\gamma$ are hyperparameters controlling the weight of each regularization constraint.

\begin{table*}[htbp]
\centering
\small
\caption{
Performance comparison on the CMU-MOSEI dataset under aligned and unaligned settings. As LLM-based methods process raw video segments, aligned results are unavailable. Best results are \colorbox{red!10}{red}, second-best are \colorbox{teal!10}{blue}. A full comparison between multimodal methods, MLLMs and classical methods is in Appendix \ref{sec:appendixB}.
}
\label{tab:cmu-mosei}
\begin{tabular}{cc|cccc|cccc}
\hline
\multirow{2}{*}{Approaches} & \multirow{2}{*}{Methods} & \multicolumn{4}{c|}{Aligned} & \multicolumn{4}{c}{Unaligned} \\ \cline{3-10} 
& & Acc & P & R & miF1 & Acc & P & R & miF1 \\ \hline
\cline{1-10}
\multirow{3}{*}{LLM-based} & GPT-4o & ---- & ---- & ---- & ----   & 0.352 & 0.583 & 0.252 & 0.196 \\
& Qwen2-VL-7B & ---- & ---- & ---- & ----   & 0.422 & 0.520 & 0.355 & 0.355 \\
& AnyGPT & ---- & ---- & ---- & ----   & 0.134 & 0.251 & 0.445 & 0.321 \\
\hline
\multirow{8}{*}{Multimodal} 
& DFG & 0.396 & 0.595 & 0.457 & 0.517 & 0.386 & 0.534 & 0.456 & 0.494 \\
& RAVEN & 0.416 & 0.588 & 0.461 & 0.517 & 0.403 & 0.633 & 0.429 & 0.511 \\
& MulT & 0.445 & 0.619 & 0.465 & 0.501 & 0.423 & 0.636 & 0.445 & 0.523 \\
& MISA & 0.430 & 0.453 & \colorbox{red!10}{\textbf{0.582}} & 0.509   & 0.398 & 0.371 & \colorbox{red!10}{\textbf{0.571}} & 0.450 \\
& MMS2S & 0.475 & 0.629 & 0.504 & 0.516   & 0.447 & 0.619 & 0.462 & 0.529 \\
& HHMPN & 0.459 & 0.602 & 0.496 & 0.556   & 0.434 & 0.591 & 0.476 & 0.528 \\
& TAILOR & 0.488 & 0.641 & 0.512 & 0.569   & 0.460 & 0.639 & 0.452 & 0.529 \\
& AMP & 0.484 & 0.643 & 0.511 & 0.569   & 0.462 & \colorbox{teal!10}{\textbf{0.642}} & 0.459 & 0.535 \\ 
& CARAT & \colorbox{teal!10}{\textbf{0.494}} & \colorbox{red!10}{\textbf{0.661}} & 0.518 & \colorbox{teal!10}{\textbf{0.581}} & \colorbox{teal!10}{\textbf{0.466}} & \colorbox{red!10}{\textbf{0.652}} & 0.466 & \colorbox{teal!10}{\textbf{0.544}} \\

\cline{2-10}
& LDDU & \colorbox{red!10}{\textbf{0.494}} & \colorbox{teal!10}{\textbf{0.647}} & \colorbox{teal!10}{\textbf{0.531}} & \colorbox{red!10}{\textbf{0.587}} & \colorbox{red!10}{\textbf{0.496}} & 0.638 & \colorbox{teal!10}{\textbf{0.543}} & \colorbox{red!10}{\textbf{0.587}} \\ 

\hline                 
\end{tabular}
\end{table*}

\section{Experiments}
\subsection{Experimental Setup}
\paragraph{Datasets and Evaluation Metrics.}
We validate the proposed method LDDU on two benchmark:  CMU-MOSEI \citep{b19} and M3ED \citep{b16}. CMU-MOSEI consists of 23,453 video segments across 250 topics. Each segment is labeled with multiple emotions, including happiness, sadness, anger, disgust, surprise, and fear. The M$^3$ED dataset, designed for dialog emotion recognition, offers greater volume and diversity compared to IEMOCAP \citep{b34} and MELD \citep{b35}. It includes 24,449 segments, capturing diverse emotional interactions with seven emotion categories: the above six emotions with neutral. Following previous work \cite{b3, b7, b24, b5}, in the experiments, we evaluate model performance using accuracy (Acc), precision (P), recall (R), and micro-F1 score (miF1).
\begin{table}[t]
\centering
\small
\caption{Performance comparison on the M$^3$ED dataset.}
\label{tab:m3ed}
\begin{tabular}{>{\centering\arraybackslash}p{1.2cm}|>{\centering\arraybackslash}p{0.9cm}>{\centering\arraybackslash}p{0.9cm}>{\centering\arraybackslash}p{0.9cm}>{\centering\arraybackslash}p{0.9cm}}
\hline
Methods & Acc & P & R & miF1 \\ \hline
MMS2S  & 0.645 & 0.813 & 0.737 & 0.773 \\
HHMPN  & 0.648 & 0.816 & 0.743 & 0.778 \\
TAILOR & 0.647 & 0.814 & 0.739 & 0.775 \\
AMP    & 0.654 & 0.819 & 0.748 & 0.782 \\
CARAT  & \colorbox{teal!10}{\textbf{0.664}} & \colorbox{teal!10}{\textbf{0.824}} & \colorbox{teal!10}{\textbf{0.755}} & \colorbox{teal!10}{\textbf{0.788}} \\ 
\hline
LDDU  & \colorbox{red!10}{\textbf{0.690}} & \colorbox{red!10}{\textbf{0.843}} & \colorbox{red!10}{\textbf{0.774}} & \colorbox{red!10}{\textbf{0.807}} \\
\hline
\end{tabular}
\end{table}

\paragraph{Baselines.} We compare the LDDU model with two types methods: traditional multimodal methods and multimodal large language model (MLLM) methods. Traditional methods include DFG \citep{b42}, RAVEN \citep{b43}, MulT \citep{b44}, MISA \citep{b23}, MMS2S \citep{b9}, HHMPN \citep{b3}, TAILOR \citep{b7}, AMP \citep{b24}, and CARAT \citep{b5}. 

Furthermore, given the significant success of MLLMs in multimodal tasks, we compare LDDU with MLLMs such as GPT-4o (\textit{gpt-4o-2024-11-20}) \citep{b57}, Qwen2-VL-7B \citep{b58}, and AnyGPT \citep{anygpt}. They respectively correspond to the open-source paradigm, closed-source paradigm, and the omni large language model (LLM). We conduct experiments using raw video clips (treated as unalgined data) from the CMU-MOSEI dataset, maintaining consistent prompts and experimental settings with the framework proposed by \citet{b59}. Details of the prompts are provided in Appendix \ref{sec:appendixA}.

In addition, we conducted a comprehensive comparison between the LDDU and existing multi-label classification (MLC) approaches including both classical methods: BR~\cite{br}, LP~\cite{lp}, CC~\cite{cc}; and single-modality methods: SGM~\cite{sgm}, LSAN~\cite{lsan}, ML-GCN~\cite{mlgcn}, please see Appendix \ref{sec:appendixB} and Table~\ref{tab:total} for full comparisons. 
\paragraph{Implementation Details.} We set $\lambda = 0.1$, $\beta = 0.8$, and $\gamma = 0.1$, with a batch size of 128. The learning rate is 2e-5 with 30 epochs. More details of all experiences are shown in Appendix~\ref{sec:appendixC}.
\subsection{Experimental Results}
\paragraph{Main Results.}In Table \ref{tab:cmu-mosei} and Table \ref{tab:m3ed}, we compare the performance of our method with various baseline approaches on the CMU-MOSEI and M$^3$ED datasets. Different from most baseline methods listed in Table 1, which use the CTC \citep{b60} module to align non-aligned datasets, LDDU performs greatly better on unaligned data without relying on the CTC module. The emotion extraction network in LDDU directly extracts modality-specific features related to the labels from the sequence data, unaffected by the varying sequence lengths across modalities.

Based on Tables \ref{tab:cmu-mosei} and  \ref{tab:m3ed}, we can draw the following observations: (1) LDDU outperforms other baseline methods on more crucial metrics such as mi-F1 and accuracy(acc) although recall and precision scores are not the highest on the CMU-MOSEI dataset. Notably, LDDU achieved balanced performance on both aligned and unaligned datasets, with unaligned's accuracy improved by 3\% and unaligned's mi-F1 increased by 4.3\%. This demonstrates that by modeling the emotional space rather than sequence features, LDDU can better capture emotion-related features. (2) LDDU also achieved significant improvements across all metrics on the M$^3$ED dataset, confirming the robustness of our model. (3) TAILOR, CAFET, and the proposed LDDU approach performed better by separating features, emphasizing the importance of considering each modality's unique contributions to emotion recognition in MMER tasks. (4) While MLLMs are excellent at video understanding, the proposed method significantly outperforms MLLMs. This maybe because their ability to capture finer-grained emotional information is limited and smaller models outperform them in this regard.

\begin{table}[t]
\centering
\small
\setlength{\tabcolsep}{3pt}
\caption{Ablation tests on the aligned CMU-MOSEI.}
\label{tab:ablation}
\begin{tabular}{>{\raggedright\arraybackslash}p{2.9cm}|>{\centering\arraybackslash}c>{\centering\arraybackslash}c>{\centering\arraybackslash}c>{\centering\arraybackslash}c}

\hline
Approach & Acc & P & R & miF1 \\ \hline
(1) w/o ESM & 0.478 & 0.663 & 0.510 & 0.577 \\
(2) w/o $\mathcal{L}_{dir}$  & 0.491 & 0.656 & 0.521 & 0.580 \\
\hline
(3) w/o $\mathcal{L}_{scl}$ & 0.483 & \colorbox{red!10}{\textbf{0.679}} & 0.498 & 0.575 \\
(4) w/o queue $\mathcal{Q}$ & 0.487 & 0.655 & 0.487 & 0.578 \\
(5) w/o variance $\mu$ & 0.483 & 0.628 & 0.536 & 0.578 \\ 
(6) w/o center $\sigma$ & 0.492 & 0.647 & 0.527 & 0.581 \\ 
\hline
(7) w/o $\mathcal{L}_{ocl}$ & 0.483 & 0.672 & 0.510 & 0.580 \\ 
(8) ow Corr(S, R) & 0.484 & 0.641 & 0.532 & 0.581 \\
(9) ow Corr(D, R) & 0.490 & 0.647 & 0.533 & 0.584 \\
(10) ow Corr(D, S) & 0.492 & 0.633 & 0.538 & 0.582 \\
\hline
(11) $\mathcal{L}_{cls}$ w/o $\hat{y}^{\mu}$ & 0.485 & 0.666 & 0.510 & 0.578 \\
(12) $\mathcal{L}_{cls}$ w/o $\hat{y}^{\sigma}$ & 0.494 & 0.622 & \colorbox{red!10}{\textbf{0.543}} & 0.580 \\
\hline
(12) LDDU & \colorbox{red!10}{\textbf{0.494}} & 0.647 & 0.531 & \colorbox{red!10}{\textbf{0.587}} \\
\hline
\end{tabular}
\end{table}

\paragraph{Ablation Study.} To elucidate the significance of each component of proposed methods, we compared LDDU against various ablation variants. As shown in Table~\ref{tab:ablation}, where "w/o" means removing, "ow" denotes only existing, "w/o ESM" denotes removing trainable feature $L$. "w/o $\sigma$, $\mu$" respectively means only consider variances or centers during CL, "w/o Corr(S,R), Corr(D,R)" denotes ignoring the calibration of (S,R) or (D,R), "$\mathcal{L}_{cls}$ w/o $\hat{y}^{\sigma}$, $\hat{y}^{\mu}$" means final classification without variance or semantic center features. We could find:

\begin{figure}[t]
    \centering
    \begin{minipage}[b]{0.233\textwidth}
        \centering
        \includegraphics[width=\textwidth]{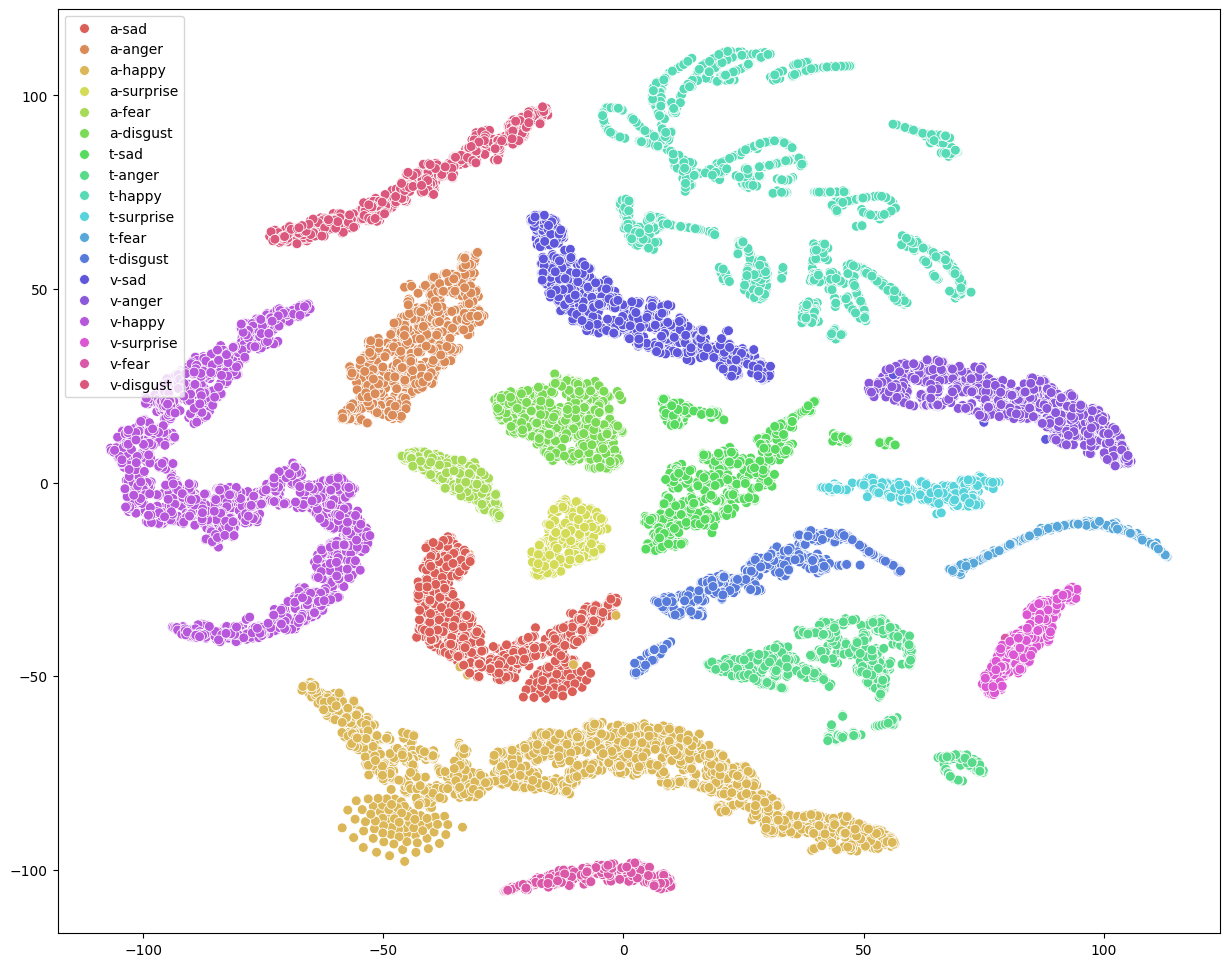}
        \caption*{(a). t-SNE with CL}  
    \end{minipage}
    \hspace{0\textwidth}  
    \begin{minipage}[b]{0.233\textwidth}
        \centering
        \includegraphics[width=\textwidth]{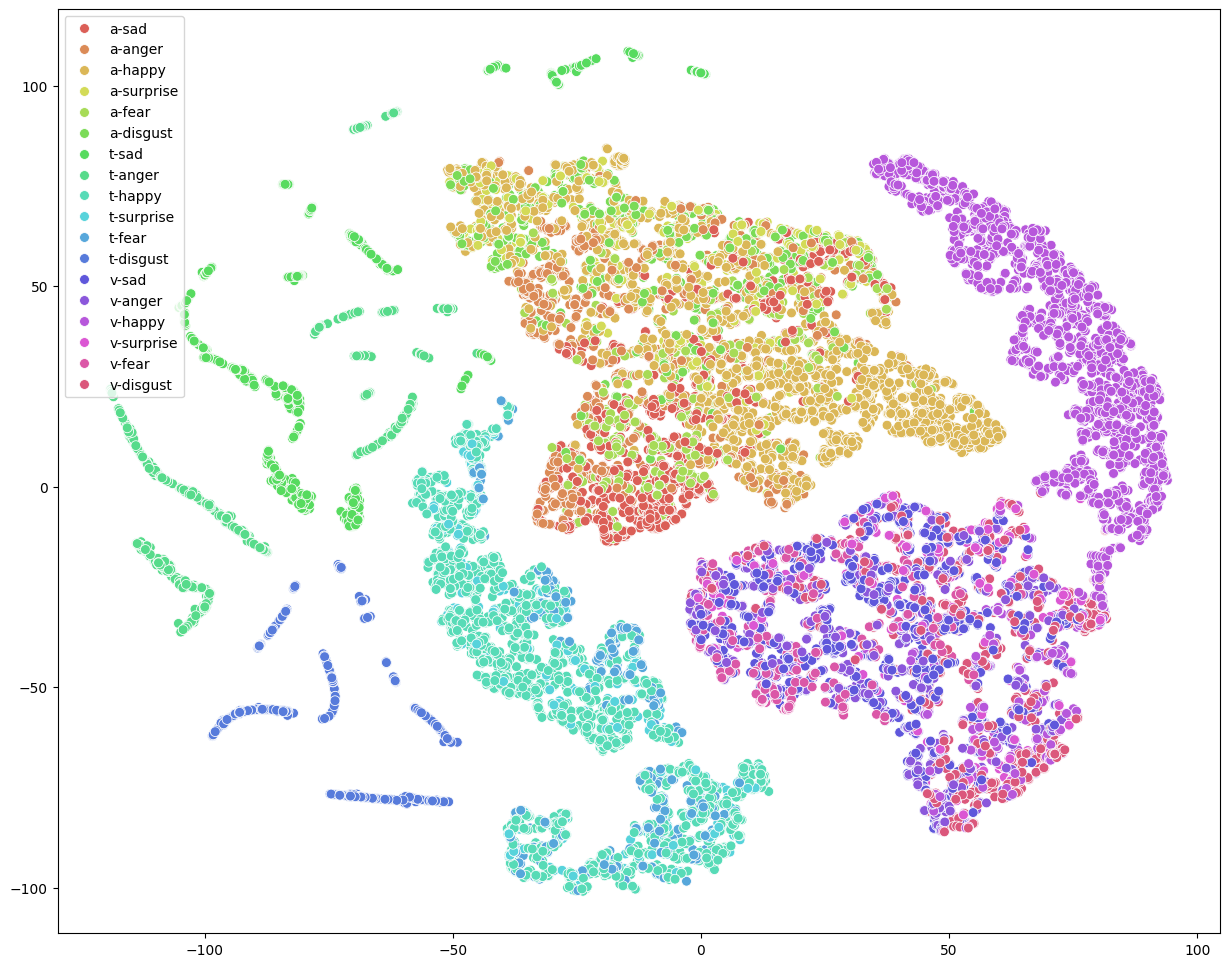}
        \caption*{(b). t-SNE without CL}  
    \end{minipage}
    \caption{The t-SNE visualization of embedding with /without CL datasets. Different colors represent different label-related modality-special features of samples.}
    \label{fig:t-sne}
\end{figure}

1) \textit{Effect of emotion space modeling (ESM)}. We replaced ESM with MLP-based attention in (1) and dropped the loss $\mathcal{L}_{dir}$ in (2). Both of them illustrated the trainable features $L$ with supervisory signals from $\mathcal{L}_{dir}$ can learn more distinguishable features of raw multimodel sequences

2) \textit{Effect of contrastive learning.} We compared LDDU with the variants without $\mathcal{L}_{scl}$ in (3). Performance degradation across metrics confirms the essential role of CL in decoupling. (4) is better than (3), which illustrates a larger batch size can enhance CL. Further, (5) and (6) demonstrates when computing similarity between distributions, both mean value and variance should be considered. 

3) \textit{Effect of uncertainty calibration.} Compared with variants without cilbration, the implementations of constraints (8, 9, 10, 12) show enhanced performance. This calibration aligned the variance with uncertainty, generating better predictions.

4) \textit{Effect of uncertainty-awared fusion.} To modeling aleatoric uncertainty, we integrated the semantic features with the distribution's regional information. (11) and (12) illustrates that both of them contributes to the final classification. 

\subsection{Further Analysis}
\paragraph{Visualization of Emotion Distribution.} To evaluate the effectiveness of Contrastive Learning (CL), we used t-SNE \citep{b61} to visualize latent emotion distributions from the CMU-MOSEI test set, excluding samples without specific emotions. As shown in Fig.~\ref{fig:t-sne}, panels (a) and (b) display distributions with and without CL, respectively. Without CL, a clear modality gap exists and intra-modality distributions lack distinctiveness. With CL, the $3 \times nl$ emotion distributions across labels and modalities are distinctly separated, enhancing their distinguishability. Consequently, LDDU leveraging CL can more effectively learn emotion distributions across modalities within the joint emotional space, with each cluster representing a specific emotion.

\begin{figure}[t]
\small
\centerline{\includegraphics[width=0.45\textwidth]{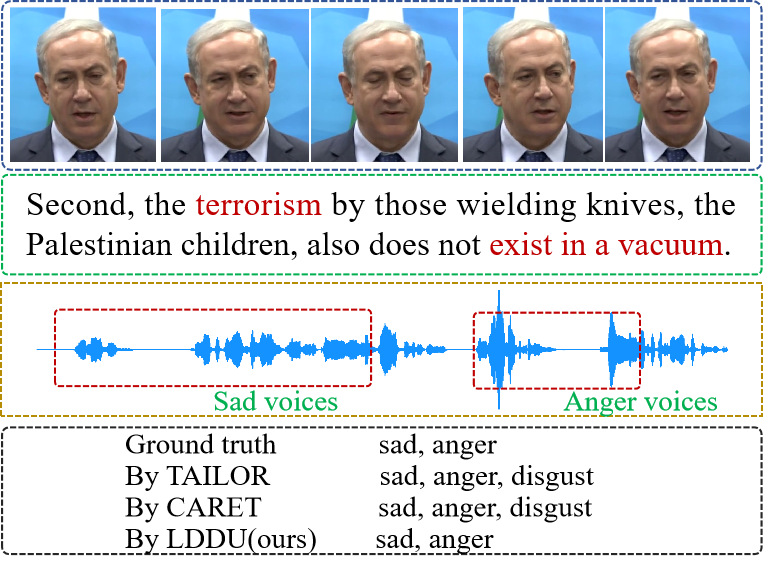}}
\caption{The case of emotion recognition by multiple methods. Visual and acoustic modalities revealed a shift from sadness to anger, while the textual modality explicitly indicated anger-related expressions. }
\label{case1}
\end{figure}

\paragraph{Case Study.}To validate LDDU's effectiveness, Figure~\ref{case1} illustrates a representative case where visual/acoustic modalities indicate a transition from sadness to anger, while textual modality explicitly signals anger. While all methods accurately detected sadness and anger, TAILOR and CARET falsely predicted disgust due to its ambiguous emotional cues in overlapping scenarios. In contrast, LDDU effectively modeled emotion-specific uncertainties through latent space Gaussian distributions (distance vector $D$'s value: sad: 0.23, anger: 0.41, disgust: 0.82). We further computed emotion correlation matrices ($M_1$–$M_3$ for methods, $M_0$ for ground truth) and measure their cosine similarities with $M_0$: LDDU achieved 96.7\% (vs. 93.3\% for TAILOR, 96.1\% for CARET), demonstrating superior capability in capturing inter-emotion dependencies. More cases are shown in Appendix~\ref{sec:appendixD}.

\section{Conclusion}
We propose LDDU, a framework that captures aleatoric uncertainty in MMER through latent emotional space probabilistic modeling. By disentangling semantic features and uncertainty using Gaussian distributions, LDDU mitigates ambiguity arising from variations in emotional intensity and overlapping emotions. Furthermore, an uncertainty-aware fusion module adaptively integrates multimodal features based on their distributional uncertainty.
Experimental results on CMU-MOSEI and M$^3$ED demonstrate that LDDU achieves state-of-the-art performance. This work pioneers probabilistic emotion space modeling, providing valuable insights into uncertainty-aware affective computing.

\section*{Limitation}
While LDDU demonstrates promising performance in MMER, several problems remain to discuss. 
LDDU models emotion uncertainty using Gaussian distributions in the latent emotion space, effectively capturing inherent ambiguity. However, it does not explicitly utilize emotion intensity labels, as provided in the CMU-MOSEI dataset (quantized into 0, 0.3, 0.6, and 1.0 levels). While this omission ensures fair comparisons with prior work (e.g., TAILOR, CARET), it also limits LDDU’s ability to precisely distinguish emotions of varying intensities. As a result, the model may be less effective in disambiguating overlapping emotions, particularly in tasks requiring fine-grained intensity differentiation. Integrating explicit intensity supervision in future iterations could further refine LDDU’s predictive capability.
\section*{Ethical Considerations}
Ethical considerations are crucial in multimodal emotion recognition research, particularly with sensitive human data like emotional expressions. In our study, we ensure that all datasets, including CMU-MOSEI and M$^3$ED, are publicly available and anonymized to protect individuals' privacy. 

While our method advances emotion recognition in areas such as human-computer interaction, we acknowledge the potential for misuse, such as manipulation or surveillance. We emphasize the responsible use of these technologies, ensuring they are deployed in contexts that respect privacy. 

Additionally, emotional expressions vary across cultures and individuals, and our model may not fully capture this diversity. We recommend expanding datasets to include a wider range of cultural contexts to avoid biases and misinterpretations. 

Finally, we commit to transparency by making our code publicly available for further scrutiny and improvement, ensuring our research aligns with ethical principles and benefits society.
\section*{Acknowledgements}
We sincerely thank all the anonymous reviewers. The work is supported by the National Natural Science Foundation of China (62206267 and 62176029), Chongqing Key Project of Technological Innovation and Application Development (CSTB2023TIAD-KPX0064), China Postdoctoral Science Foundation Funded Project (2024M763867).
\bibliography{custom}

\newpage

\appendix
\section{Appendix}
\subsection{Implementation Details}
\label{sec:appendixC}
We set $\lambda = 0.1$, $\beta = 0.8$, and $\gamma = 0.1$, with a batch size of 128. For the uni-modal extraction network, each Transformer consists of 3 layers ($l_a = l_v = l_t = 3$). The hidden dimensions are 256 for feature $Y$ and 128 for feature $Z$. The latent emotion distribution has a dimension of 64 for both the distribution centers and variance vectors. The contrastive learning queue $\mathcal{Q}$ is sized at 8192. The number of labels ($q$) is 6 for CMU-MOSEI and 7 for M$^3$ED. We optimize all model parameters using the Adam optimizer \citep{b56} with a learning rate of $2 \times 10^{-5}$ and a cosine decay schedule with a warm-up rate of 0.1. All experiments are conducted on a single GTX A6000 GPU using grid search.

\subsection{More Compared Baselines}
\label{sec:appendixB}
Despite the advancements in LLM-based and multimodal methods, we conducted a comprehensive and comparative analysis between the LDDU model and existing multi-label classification (MLC) approaches. This comparison includes both classical methods (BR \cite{br}, LP \cite{lp}, CC \cite{cc}) and single-modality methods (SGM \cite{sgm}, LSAN \cite{lsan}, ML-GCN \cite{mlgcn}). The experimental results are presented in Table \ref{tab:total}.

\begin{table*}[htbp]
\centering
\caption{
Performance comparison on the CMU-MOSEI dataset under aligned and unaligned settings. As LLM-based methods process raw video segments, aligned results are unavailable. Best results are \colorbox{red!10}{red}, second-best are \colorbox{teal!10}{blue}.
}
\label{tab:total}
\begin{tabular}{cc|>{\centering\arraybackslash}p{1.0cm}>{\centering\arraybackslash}p{1.0cm}>{\centering\arraybackslash}p{1.0cm}>{\centering\arraybackslash}p{1.0cm}|>{\centering\arraybackslash}p{1.0cm}>{\centering\arraybackslash}p{1.0cm}>{\centering\arraybackslash}p{1.0cm}>{\centering\arraybackslash}p{1.0cm}}

\hline
\multirow{2}{*}{Approaches} & \multirow{2}{*}{Methods} & \multicolumn{4}{c|}{Aligned} & \multicolumn{4}{c}{Unaligned} \\ \cline{3-10} 
& & Acc & P & R & miF1 & Acc & P & R & miF1 \\ \hline
\cline{1-10}
\multirow{3}{*}{LLM-based} & GPT-4o & ---- & ---- & ---- & ----   & 0.352 & 0.583 & 0.252 & 0.196 \\
& Qwen2-VL-7B & ---- & ---- & ---- & ----   & 0.422 & 0.520 & 0.355 & 0.355 \\
& AnyGPT & ---- & ---- & ---- & ----   & 0.134 & 0.251 & 0.445 & 0.321 \\
\hline
\multirow{3}{*}{Classical} & BR & 0.222 & 0.309 & 0.515 & 0.386 & 0.233 & 0.321 & 0.545 & 0.404 \\
& LP & 0.159 & 0.231 & 0.377 & 0.286 & 0.185 & 0.252 & 0.427 & 0.317 \\
& CC & 0.225 & 0.306 & 0.523 & 0.386 & 0.235 & 0.320 & 0.550 & 0.404 \\
\hline
\multirow{3}{*}{Deep-based} & SGM & 0.455 & 0.595 & 0.467 & 0.523 & 0.449 & 0.584 & 0.476 & 0.524 \\
& LSAN & 0.393 & 0.550 & 0.459 & 0.501 & 0.403 & 0.582 & 0.460 & 0.514 \\
& ML-GCN & 0.411 & 0.546 & 0.476 & 0.509 & 0.437 & 0.573 & 0.482 & 0.524 \\
\hline
\multirow{8}{*}{Multimodal} 
& DFG & 0.396 & 0.595 & 0.457 & 0.517 & 0.386 & 0.534 & 0.456 & 0.494 \\
& RAVEN & 0.416 & 0.588 & 0.461 & 0.517 & 0.403 & 0.633 & 0.429 & 0.511 \\
& MulT & 0.445 & 0.619 & 0.465 & 0.501 & 0.423 & 0.636 & 0.445 & 0.523 \\
& MISA & 0.430 & 0.453 & \colorbox{red!10}{\textbf{0.582}} & 0.509   & 0.398 & 0.371 & \colorbox{red!10}{\textbf{0.571}} & 0.450 \\
& MMS2S & 0.475 & 0.629 & 0.504 & 0.516   & 0.447 & 0.619 & 0.462 & 0.529 \\
& HHMPN & 0.459 & 0.602 & 0.496 & 0.556   & 0.434 & 0.591 & 0.476 & 0.528 \\
& TAILOR & 0.488 & 0.641 & 0.512 & 0.569   & 0.460 & 0.639 & 0.452 & 0.529 \\
& AMP & 0.484 & 0.643 & 0.511 & 0.569   & 0.462 & \colorbox{teal!10}{\textbf{0.642}} & 0.459 & 0.535 \\ 
& CARAT & \colorbox{teal!10}{\textbf{0.494}} & \colorbox{red!10}{\textbf{0.661}} & 0.518 & \colorbox{teal!10}{\textbf{0.581}} & \colorbox{teal!10}{\textbf{0.466}} & \colorbox{red!10}{\textbf{0.652}} & 0.466 & \colorbox{teal!10}{\textbf{0.544}} \\

\cline{2-10}
& LDDU & \colorbox{red!10}{\textbf{0.494}} & \colorbox{teal!10}{\textbf{0.647}} & \colorbox{teal!10}{\textbf{0.531}} & \colorbox{red!10}{\textbf{0.587}} & \colorbox{red!10}{\textbf{0.496}} & 0.638 & \colorbox{teal!10}{\textbf{0.543}} & \colorbox{red!10}{\textbf{0.587}} \\ 

\hline                 
\end{tabular}
\end{table*}

\subsection{Prompts of MLLM}
\label{sec:appendixA}
\begin{figure*}
\centerline{\includegraphics[width=1\textwidth]{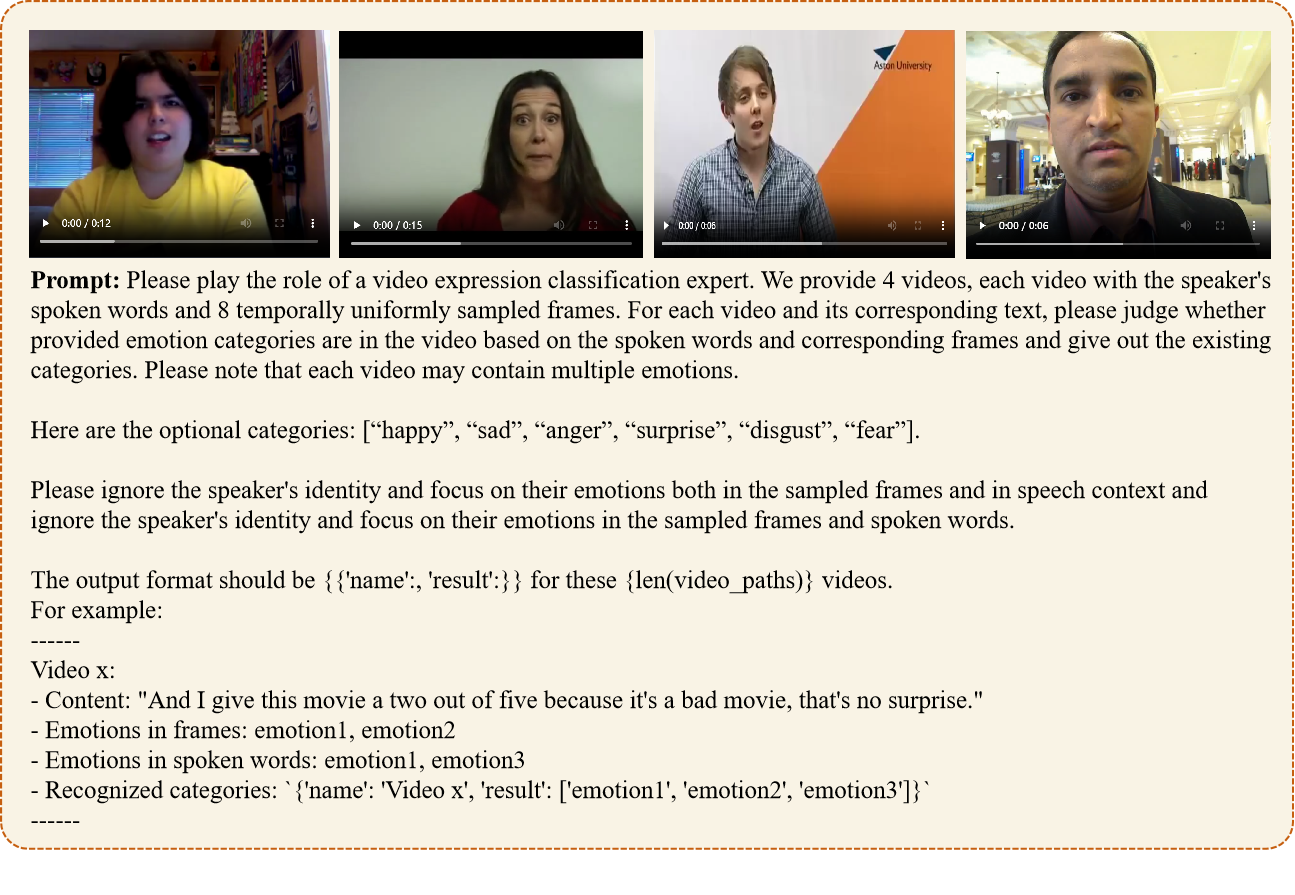}} 
\caption{Prompts of MLLMs.}
\label{gpt-case}
\end{figure*}

In this study, we evaluated three multimodal models (GPT-4o, Qwen2-VL-7B, and AnyGPT), using video clips with an average duration of 7–8 seconds. GPT-4o and Qwen2-VL-7B exhibit strong visual understanding capabilities, representing closed-source and open-source multimodal large language models (MLLMs), respectively. AnyGPT is a versatile any-to-any MLLM capable of processing images, text, and audio. Since all these MLLMs adopt end-to-end architectures, we ensured computational efficiency and consistency by uniformly sampling 8 frames per video clip as input for inference. The specific prompts designed for each model, including task descriptions and format requirements, are detailed in Figure \ref{gpt-case}.

\subsection{Detailed Info of Uncertainty Caliration}
\label{sec:appendixE}
To enhance readers' understanding of aleatoric uncertainty and uncertainty correction, we provide additional supplementary materials.

\subsubsection{Aleatoric Uncertainty in MMER}
Aleatoric uncertainty refers to the inherent variability or noise present in the data, arising from factors beyond the model’s control. In the context of emotion recognition, it stems from factors such as variations in emotional intensity, individual differences, and the blending of multiple emotions. This form of uncertainty is intrinsic to the data itself.

In multimodal emotion recognition (MMER), aleatoric uncertainty becomes particularly evident when the same emotion is expressed by different individuals. For example, a person may express happiness through a broad smile (visual modality) but with a neutral tone of voice (audio modality), reflecting differences in emotional intensity and expression. These inconsistencies can introduce conflicting cues that complicate the emotion recognition process. Furthermore, datasets like CMU-MOSEI also contain varying levels of emotion intensity, further contributing to aleatoric uncertainty.

This type of uncertainty is not confined to emotion recognition alone. In computer vision (CV), it can manifest as blurry faces or imprecise object localization, introducing uncertainty in tasks like object detection. In natural language processing (NLP), aleatoric uncertainty arises from ambiguities in language, where word meanings can shift based on contextual factors. In all these scenarios, probabilistic models are employed to capture and account for such inherent uncertainty, thereby enhancing the robustness of systems in diverse, real-world environments.

\subsubsection{Uncertainty Calibration}
\paragraph{Uncertainty Calibration.} Uncertainty Calibration refers to the process of adjusting model predictions to more accurately reflect the true uncertainty associated with them. In machine learning and deep learning, models often provide predictions accompanied by an associated uncertainty; however, these predictions are not always well-calibrated. In other words, the model may exhibit excessive confidence in certain predictions, even when the true uncertainty is high, or it may fail to properly estimate its own uncertainty.

The primary objective of uncertainty calibration is to align the predicted uncertainty with the actual likelihood of a prediction being correct. In practical terms, this means that if a model is 90\% confident in its prediction, it should be correct approximately 90\% of the time over a large number of predictions. This calibration process is particularly critical in domains such as emotion recognition, medical diagnosis, and autonomous driving, where accurate uncertainty estimates are essential for reliable decision-making. Several methods can be employed for uncertainty calibration, including temperature scaling, Platt scaling, and Bayesian approaches.

\paragraph{Ordinality Constraint.} Ordinality Constraint refers to a form of uncertainty calibration that is based on the ranking of classes. This method assumes that the relationship between classes or labels follows a natural ordinal structure, where labels have an inherent order. For instance, in sentiment analysis, labels such as "very negative," "negative," "neutral," "positive," and "very positive" exhibit a natural progression from negative to positive sentiment. Ordinality constraints ensure that the model’s predicted probabilities reflect this ranking, adjusting the output so that predictions align with the ordered nature of the classes.

In our proposed approach, the ordinality constraint is applied to rank the uncertainty of predictions across different labels. By incorporating this constraint, we ensure that the model not only outputs probabilities but also ranks the classes in a manner that respects their inherent order.

\subsubsection{Uncertainty Caliration in LDDU}
Since networks learning variance and mean vectors share similar structures, variance and mean tend to converge and surface feature space collapse without constraints. The key is to ensure that variance vectors accurately reflect uncertainty level. We introduce an ordinality (ranking) constraint \citep{b29} to solve this problem.
As shown in Equation \ref{eq:21}, ordinality constraint requires predicted confidence $\kappa$ should correspond to the probability $\mathcal{P}$ of correct prediction. In our approach, the variance $\sigma_i = (\sigma_i^v, \sigma_i^a, \sigma_i^t)$ and the prediction error $d(\hat{y}_i^{dir}, y_i)$ from the Info Classifier jointly represent the sample's confidence. The main challenge is establishing reliable proxy features for $\mathcal{P}$. Inspired by CRL \citep{b26}, we use the proportion of samples $r_i$ correctly predicted by the Info Classifier during the SGD \citep{b52} process as a proxy for $\mathcal{P}$. Empirical findings from \citet{b53} and \citet{b54} support our hypothesis: frequently forgotten samples are harder to classify, while easier samples are learned earlier in training.

When the sample contain high uncertainty, the latent distribution variance $\sigma_i$ and the prediction error $d_i=d(\hat{y}_i^{dir}, y_i)$ tend to be large, while $r_i$ tend to be small. Conversely, when the uncertainty is small, these features are reversed. Therefore, the ordinality constraint is:
\begin{align}
max\ Corr(rk(\frac{1}{||\sigma_i||_2},\frac{1}{||\sigma_j||_2}),\ rk(r_i,r_j))   \\
argmax\ Corr(rk(1-d_i, 1-d_j),\ rk(r_i,r_j))  
\end{align}
where $Corr$ represents correlation and $rk$ demotes ranking.  In this paper, we impose ordinality constraints based on soft-ranking\citep{b25, b55}. While \citep{b25} uses KL divergence to measure mismatching of softmax distributions and \citep{b55} applies softmax cross-entropy for ordinal regression, our method employs bidirectional KL divergence to assess mismatching between the softmax distributions 

For a batch of size $s_B$, we compute the variance norm $S$, distance vector $D$, and proxy vector $R$ for each sample:
\begin{align}
S=[&\frac{1}{||\sigma_1||_2},\frac{1}{||\sigma_2||_2},..,\frac{1}{||\sigma_{s_B}||_2}] \\
D=[&1-d_1,1-d_2,..,1-d_{s_B}] \\
&R=[r_1,r_2,..,r_{s_B}]    
\end{align}

Inspired by \citep{b25, b55}, we impose ordinality constraints based on soft-ranking. While \citep{b25} uses KL divergence to measure mismatching of softmax distributions and \citep{b55} applies softmax cross-entropy for ordinal regression, our method employs bidirectional KL divergence to assess mismatching between the softmax distributions of pairs $(S,R)$ and $(D, R)$. Consequently, ordinality calibration loss $\mathcal{L}_{ocl}$can be calculated as follows:
\begin{align}
\mathcal{L}_{ocl} = & KL(P_D||P_R) + KL(P_R||P_D) \notag \\ &+ KL(P_S||P_R) + KL(P_R||P_S)
\end{align}
where $P_D$, $P_R$, and $P_S$ represent the softmax distributions of features $S$, $R$, and $D$, respectively.

\begin{figure}[t]
\centering
\includegraphics[width=0.45\textwidth]{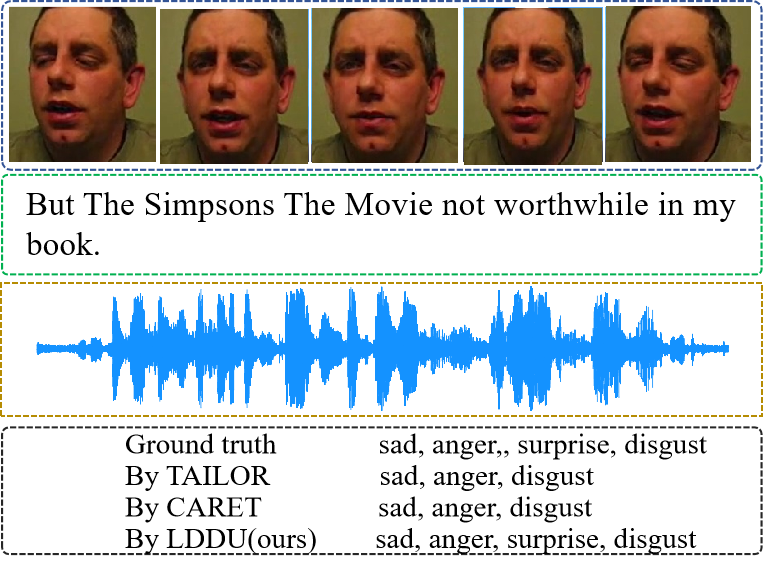}
\caption{The case of emotion recognition by multiple methods.}
\label{case2}
\end{figure}

In summary, the total training loss is as follows:
\begin{equation}
    \mathcal{L}_{total} = \mathcal{L}_{cls} + \lambda \mathcal{L}_{ocl} + \beta \mathcal{L}_{scl} + \gamma \mathcal{L}_{dir}
\end{equation}
where $\lambda$, $\beta$, and $\gamma$ are hyperparameters controlling the strength of each regularization constraint.

\subsection{More Cases for Case Study}
\label{sec:appendixD}

Another case is shown in  Figure~\ref{case2}.

\end{document}